\RequirePackage{fix-cm}

\documentclass[twocolumn]{svjour3}          % twocolumn
\smartqed  % flush right qed marks, e.g. at end of proof
\usepackage{graphicx}
\usepackage[export]{adjustbox}
\usepackage[ruled,vlined]{algorithm2e}
\usepackage{multirow}
\usepackage[noend]{algpseudocode}
\usepackage{subcaption}
\usepackage{xcolor}
\usepackage{hyphenat}
\usepackage{microtype}

\usepackage{comment}
\usepackage{multicol}
\usepackage{xcolor}
\usepackage{enumitem}
\usepackage{amsmath}
\usepackage{tocloft}
\usepackage{amssymb}
\usepackage{array}
\usepackage{adjustbox}
\usepackage{tabularx}
\usepackage{graphicx}
\usepackage{booktabs}
\usepackage{float} 
\usepackage[utf8]{inputenc}

\begin{document}
	
	\title{Enhancing Few-Shot Image Classification through Learnable Multi-Scale Embedding and Attention Mechanisms}
	
	%\titlerunning{Short form of title}        % if too long for running head
	
	\author{Fatemeh Askari\textsuperscript{1} \and
		Amirreza Fateh\textsuperscript{1} \and
		Mohammad Reza Mohammadi\textsuperscript{1, *}}
	
	\institute{
		\textsuperscript{1}School of Computer Engineering, Iran University of Science and Technology (IUST), Tehran, Iran \\
		\email{mrmohammadi@iust.ac.ir} \\
		\textsuperscript{*}\emph{Corresponding author}
	}
	%\date{Received: date / Accepted: date}
	% The correct dates will be entered by the editor
	\maketitle
	\begin{abstract}
		{ . }
            \newline            
            In the context of few-shot classification, the goal is to train a classifier using a limited number of samples while maintaining satisfactory performance. However, traditional metric-based methods exhibit certain limitations in achieving this objective. These methods typically rely on a single distance value between the query feature and support feature, thereby overlooking the contribution of shallow features. To overcome this challenge, we propose a novel approach in this paper. Our approach involves utilizing a multi-output embedding network that maps samples into distinct feature spaces. The proposed method extracts feature vectors at different stages, enabling the model to capture both global and abstract features. By utilizing these diverse feature spaces, our model enhances its performance. Moreover, employing a self-attention mechanism improves the refinement of features at each stage, leading to even more robust representations and improved overall performance. Furthermore, assigning learnable weights to each stage significantly improved performance and results. We conducted comprehensive evaluations on the MiniImageNet and FC100 datasets, specifically in the 5-way 1-shot and 5-way 5-shot scenarios. Additionally, we performed cross-domain tasks across eight benchmark datasets, achieving high accuracy in the testing domains. These evaluations demonstrate the efficacy of our proposed method in comparison to state-of-the-art approaches.
\color{purple}{https://github.com/FatemehAskari/MSENet}\color{black}

  
		% \keywords{...}
		
		% \PACS{PACS code1 \and PACS code2 \and more}
		% \subclass{MSC code1 \and MSC code2 \and more}
	\end{abstract}

\section{Introduction}
\label{intro}
In recent years, deep learning has led to remarkable progress in the field of image recognition, significantly surpassing traditional computer vision algorithms \cite{bib1,fateh2024advancing}. However, the success of deep learning models heavily relies on the availability of large datasets. When the available data is insufficient, models struggle to optimize their parameters effectively, leading to overfitting and ultimately hindering performance \cite{bib2}. This challenge is particularly pronounced in contexts where data labeling is time-consuming and costly, such as in medical imaging or rare object classification \cite{tian2024survey,sun2024klsanet,rezvani2024fusionlungnet}. Therefore, the development of models capable of achieving acceptable performance with limited samples is critical \cite{bib7}.

Data augmentation is one technique employed to mitigate the impact of limited labeled data \cite{wang2023data}. However, traditional augmentation methods, such as rotation and noise addition, often fail to provide substantial new information, thereby limiting their effectiveness in preventing overfitting \cite{bib9}. Another approach, transfer learning \cite{bib10}, involves transferring knowledge from a source domain to a target domain by freezing shallow network layers while fine-tuning deeper layers. Yet, this method may struggle when the target domain significantly differs from the source \cite{bib11}.

To address these limitations, meta-learning has emerged as a promising solution \cite{li2023novel,fateh2024msdnet}. By leveraging prior learning experiences, meta-learning models can generalize across diverse tasks and rapidly adapt to new problem domains \cite{yang2024meta,bib14}. The primary approaches in meta-learning include model-based, optimization-based, and metric-based methods \cite{bib15}. While model-based methods focus on architecture adjustments, optimization-based methods enhance learning through episodic training \cite{bib17,bib18}. Metric-based methods, however, learn a distance metric to measure sample similarity, ensuring that samples from the same class exhibit small distances \cite{liu2024few,bib20}.

Despite their benefits, metric-based approaches typically rely on a single embedding space, which limits their ability to leverage the rich information from different feature representations. Recent advancements, such as the multi-distance metric network proposed by Gao et al. \cite{bib21}, suggest that utilizing multiple embedding spaces can enhance model performance by capturing both global and abstract features. Furthermore, the utilization of attention mechanisms and Transformers has significantly increased in recent years. For instance, methods like the SetFeat extractor introduced by Afrasiyabi et al. \cite{bib22} emphasize the importance of rich feature representations through self-attention mechanisms. These advancements highlight the growing recognition of the capabilities of attention mechanisms in enhancing accuracy and efficiency across various machine learning and computer vision tasks.

The ability to achieve high performance with limited labeled data highlights the practical value of our proposed approach, especially in scenarios where data collection and labeling are expensive or challenging. For instance, in the medical field, our model can assist in diagnosing rare diseases with only a small number of annotated samples. In industrial applications, it can support tasks such as anomaly detection or rare object classification in manufacturing lines. Similarly, in agriculture, it can facilitate the classification of plant species or pest detection with minimal labeled data.

This research aims to propose a novel few-shot classification model that integrates various innovative components to enhance performance, particularly in scenarios where labeled data is scarce. Our model employs ResNet18 as a feature extractor, extracting feature maps from multiple stages to facilitate multi-scale representation. We introduce learnable parameter weights at each stage and incorporate self-attention mechanisms to enrich the feature space. Through comprehensive evaluations on the MiniImageNet and FC100 datasets, we demonstrate the effectiveness of our approach.

Our contributions can be summarized as follows:
\begin{itemize}[label=$\bullet$]
\item We extract five feature maps from the backbone to capture both global and task-specific features.
\item We employ a self-attention mechanism for each feature map to capture more valuable information.
\item We incorporate learnable weights at each stage to enhance the model's flexibility.
\item We propose a novel few-shot classification technique that significantly improves accuracy on the MiniImageNet and FC100 datasets.
\end{itemize}

\section{Related Works}
\label{sec:1}
In this section, we discuss related work on some approaches in meta-learning.
\vspace{-\baselineskip}
\subsection*{\textbf{Model-based:}}
\vspace{-\baselineskip}
Cai et al. \cite{bib23} proposed Memory Matching Networks (MM-Net) for one-shot image recognition, which is based on the principles of Matching Networks \cite{bib24}. MM-Net combines Convolutional Neural Networks with memory modules to leverage knowledge from a set of labeled images. It employs a contextual learner to predict CNN parameters for unlabeled images. Munkhdalai et al. \cite{bib25}  proposed model, called MetaNet, consists of two main components: a base learner operating in the task space and a meta learner operating in the meta space. By leveraging meta information, MetaNet can dynamically adjust its weights to recognize new concepts in the input task. Garnelo et al. \cite{bib26} introduces a model called Conditional Neural Processes (CNPs), which combines deep neural networks with Bayesian methods. CNPs are capable of making accurate predictions after observing only a few training data points, while also being able to handle complex functions and large datasets. The disadvantage of model-based approaches is that they are computationally expensive and require significant computational resources.
\vspace{-\baselineskip}
\subsection*{\textbf{Optimization-based:}}
\vspace{-\baselineskip}
Finn et al. \cite{bib27} proposed a model-agnostic meta-learning (MAML) algorithm for fast adaptation of deep networks. The algorithm involves meta-training the model on various tasks using gradient descent to optimize its initial parameters. In the meta-testing phase, the model's performance is evaluated on new tasks sampled from a task distribution. Through gradient-based adaptation, the model fine-tunes its parameters using a small amount of data from each new task. Sun et al. \cite{bib28} proposed a novel method called Meta-Transfer Learning (MTL). MTL combines transfer learning and meta-learning to improve the convergence and generalization of deep neural networks in low-data scenarios. It introduces scaling and shifting operations to transfer knowledge across tasks. Experimental results demonstrate the effectiveness of MTL in various few-shot learning benchmarks.The disadvantage of  optimization-based approaches is that they are susceptible to issues such as getting stuck in saturation points and sensitivity to zero-gradient problems. These issues can hinder the optimization process and affect the overall performance of the method.
% \vspace{-\baselineskip}
% \vspace{-\baselineskip}
\subsection*{\textbf{Metric-based:}}
\vspace{-\baselineskip}
Koch et al. \cite{bib29} proposed a Siamese network that utilizes the VGG network as an extractor. They feed two pairs of images into the shared-weight convolutional network, and the network outputs a numerical value between 0 and 1, representing the similarity between the two images. Vinyals et al. \cite{bib24} proposed a matching network that computes the probability distribution over labels using an attention kernel. The attention kernel calculates the cosine similarity between the support set of embedded vectors and the query. It then normalizes the similarity using the softmax formula. Snell et al. \cite{bib30} proposed the Prototypical network, where each class in the support set is represented by a prototype, defined as the mean of the embedded vectors belonging to that class. The similarity between the query image's embedded vector and the prototypes of each class is determined using the Euclidean distance. This enables the classification of query images into their respective classes. Sung et al. \cite{bib31} proposed the Relation Network, which does not rely on a separate distance function. Instead, it connects the representations of the support set and the query directly within the neural network architecture, allowing the network to learn the similarity measure. Previous few-shot image classification methods commonly used four-layer convolutional networks as backbones. However utilization of pre-trained networks such as ResNet12 and ResNet18 has become much more popular nowdays. However, calculating similarities and differences to a single feature vector is not sufficient. Gao et al. \cite{bib21} proposed a model called MDM-Net for few-shot learning. The MDM-Net maps input samples into four different feature spaces using a multi-output embedding network. Additionally, they introduced a task-adaptive margin to adjust the distance between different sample pairs. Transformers and attention mechanisms have emerged as state-of-the-art solutions in few-shot image classification, surpassing traditional CNN-based approaches. While CNNs have served as reliable feature extractors, their limitations, such as a restricted receptive field and parameter inefficiency, make them less effective in capturing complex patterns. In contrast, Transformers excel by capturing long-range dependencies, modeling non-local relationships, and efficiently parallelizing computations. Additionally, they offer enhanced interpretability by identifying key regions or features in the input data, providing insights into the model's decision-making process. Recent advancements, such as the introduction of MDM-Net, demonstrate the superiority of Transformers and attention mechanisms in few-shot learning, combining their strengths to address the unique challenges of limited-data scenarios.

Wang et al. \cite{bib32} propose a unified Query-Support Transformer (QSFormer) model for few-shot learning. The QSFormer model addresses the challenges of consistent image representations in both support and query sets, as well as effective metric learning between these sets. It consists of a sampleFormer branch that captures sample relationships and conducts metric learning using Transformer encoders, decoders, and cross-attention mechanisms. Additionally, a local patch Transformer (patchFormer) module is incorporated to extract structural representations from local image patches. The proposed model also introduces a Cross-scale Interactive Feature Extractor (CIFE) as an effective backbone module for extracting and fusing multi-scale CNN features. The QSFormer model demonstrates superior performance compared to existing methods in few-shot learning.  
Ran et al. \cite{bib32} propose a novel deep transformer and few-shot learning (DT-FSL) framework for hyperspectral image classification. The framework aims to achieve fine-grained classification using only a few-shot instances. By incorporating spatial attention and spectral query modules, the framework captures the relationships between non-local spatial samples and reduces class uncertainty. The network is trained using episodes and task-based learning strategies to enhance its modeling capability. Additionally, domain adaptation techniques are employed to reduce inter-domain distribution variation and achieve distribution alignment. Cheng et al. \cite{bib79} proposed a Class-Aware Patch Embedding Adaptation (CPEA) method for few-shot image classification, leveraging Vision Transformers (ViTs) pre-trained with Masked Image Modeling to generate semantically meaningful patch embeddings. They introduced class-aware embeddings to adapt patch embeddings, enabling class-relevant comparisons without explicit localization or alignment mechanisms, achieving state-of-the-art performance.

\section{Proposed method}
\subsection{Problem definition}

\begin{figure*}[h]
	\centering
	\includegraphics[width=0.9\textwidth]{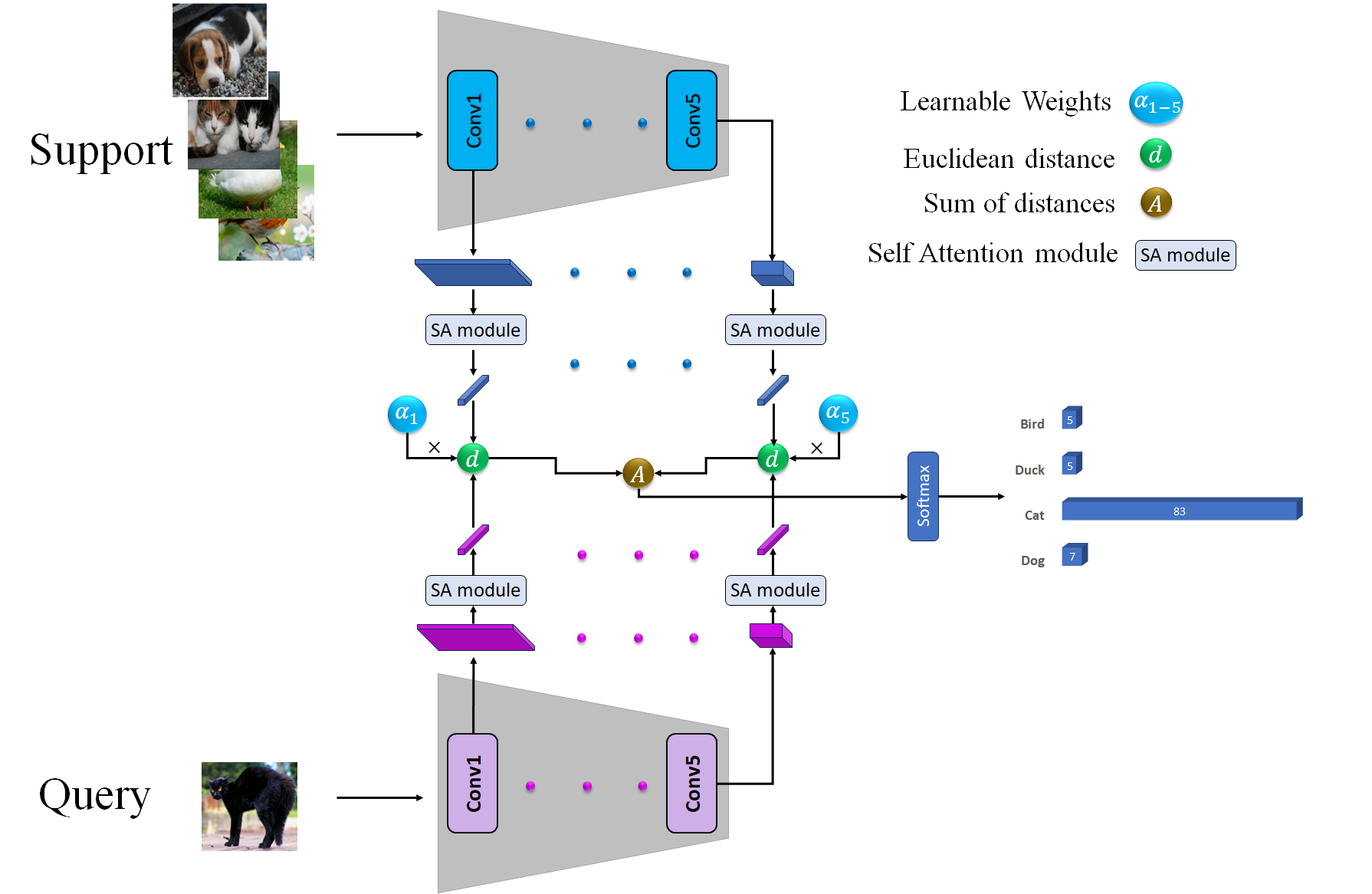}
	\caption{The overview architecture of the proposed model}
	\label{3}
\end{figure*}

The goal of few-shot classification is to classify a unseen sample. We have two datasets, $D_{train}$ and $D_{test}$, each associated with corresponding class sets $C_{train}$ and $C_{test}$, respectively. It is important the class sets $C_{train}$ and $C_{test}$ are disjoint, meaning they have no elements in common. Formally, we can express this as $C_{train} \cap C_{test} = \emptyset$. Each training episode consists of a support set $S$ and a query set $Q$. The support set $S$ comprises $K$ examples for each of $N$ distinct classes, denoted as $X_i^S=\left\{\left(x_{i j}^s, C=i\right)\right\}_{i=1}^K$, where $x_{ij}^s$ represents the $j$th example belonging to class $i$. The query set $Q$ contains $X^q=\left\{x_i^q\right\}_{i=1}^{n_q}$. The objective of the model is to leverage the support set $S$ to correctly classify the query example $x_i^q$. In other words, the model is trained to predict the class label of the query image based on the set of supporting examples and their class affiliations provided in $S$. During training, each episode consists of a random sample drawn from the training dataset $D_{train}$. The objective of the model is to learn to extract feature representations from the examples in the support set $S$, such that the distance between the feature vector of the query image $x_i^q$ and the feature vectors of the support examples $x_{ij}^s$ can be effectively measured. Specifically, the model learns to extract discriminative feature vectors from the support examples, which can then be used to classify the query image based on its proximity to the support set features. This training paradigm encourages the model to rapidly adapt its feature extraction and classification capabilities from the limited support data to accurately predict the class label of the query instance. Similarly, during evaluation, the performance of the trained model is assessed on the held-out test dataset $D_{test}$. In this phase, the model extracts feature vectors for each example in the test set, leveraging the knowledge and feature extraction capabilities it learned during the training episodes on the $D_{train}$ dataset.

\subsection{Overview}

The overall architecture of our model is illustrated in Figure~\ref{3}. Our proposed approach incorporates several key components designed to enhance performance. At the core is a robust backbone architecture that enables the extraction of feature maps across diverse spatial scales. Additionally, we have integrated an attention module to further refine the feature extraction process. Underpinning our framework is a distance metric that facilitates effective similarity computation between inputs. Moreover, we have incorporated learnable weights to capture the relative significance of each extracted feature map.
\begin{figure*}[h]
	\centering
	\includegraphics[width=0.8\textwidth]{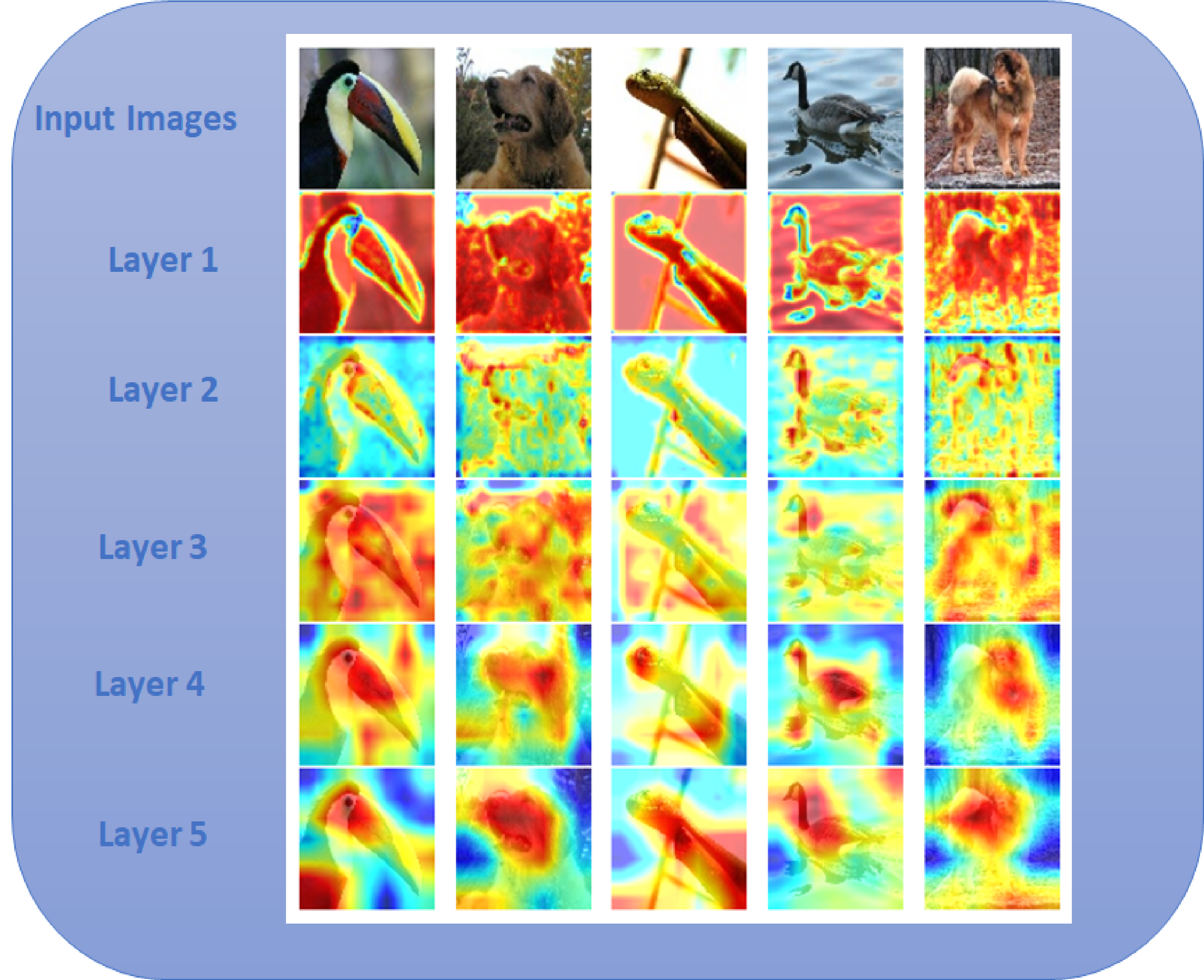}
	\caption{Visualization of feature maps from the five convolutional stages of ResNet-18, illustrating the progression from low-level features in shallow layers to high-level semantic features in deeper layers, critical for accurate classification.}
	\label{visualize}
\end{figure*}
\subsubsection{\normalfont Backbone}
We utilized a pre-trained ResNet-18 network with initial weights from the ImageNet dataset. We removed the last fully-connected layer and fine-tuned the network on our specific dataset. To leverage multi-level feature maps, we proposed a multi-output embedding approach where we extracted feature maps at the end of each of the 5 convolutional blocks in the ResNet-18 architecture. This allowed us to capture feature representations at multiple scales and resolutions. As illustrated in Figure~\ref{visualize}, the deeper layers of the ResNet-18 architecture play a more significant role in classification tasks compared to the shallow layers. While shallow layers capture low-level features such as edges and textures, the deeper layers focus on abstract and high-level semantic features that are crucial for distinguishing between classes. In our approach, we utilized the feature maps from all five stages of the ResNet-18 architecture to capture multi-scale representations. However, the deeper layers contribute more prominently to the final classification, as they extract the high-level semantic information critical for accurate class differentiation.
Each sample in the support set, denoted as $x_{ij}^s$, is mapped to five different feature spaces like $f_{ij}^s$ as shown in Equation \ref{eq_1}:
\begin{equation}
f_{i{j}}^{s} = \{ f_{i{j}}^{Conv1-s},f_{i{j}}^{Conv2-s},...,f_{i{j}}^{Conv5-s} \}
\label{eq_1}
\end{equation}
Similarly, each query sample, denoted as $x_i^q$, is mapped to the following feature space (equation \ref{eq_2}):
\begin{equation}
f_{i{j}}^{q} = \{ f_{i{j}}^{Conv1-q},f_{i{j}}^{Conv2-q},...,f_{i{j}}^{Conv5-q} \}
\label{eq_2}
\end{equation}
In our proposed approach, we employ multiple convolutional layers $(Conv1, Conv2, Conv3, Conv4, Conv5)$ of the ResNet-18 network, which act as feature extractors to transform raw input images into meaningful and structured feature representations. A feature extractor refers to a mechanism in a neural network that automatically identifies and extracts important patterns or attributes (e.g., edges, textures, shapes, or semantic structures) from raw data. By leveraging these hierarchical feature maps from both the support and query samples, our method generates a robust, multi-level representation of the input images. These extracted features provide a compact and discriminative description of the images, facilitating effective similarity computation and accurate classification, which are crucial for our task.

\begin{figure*}[h]
	\centering
	\includegraphics[width=0.8\textwidth]{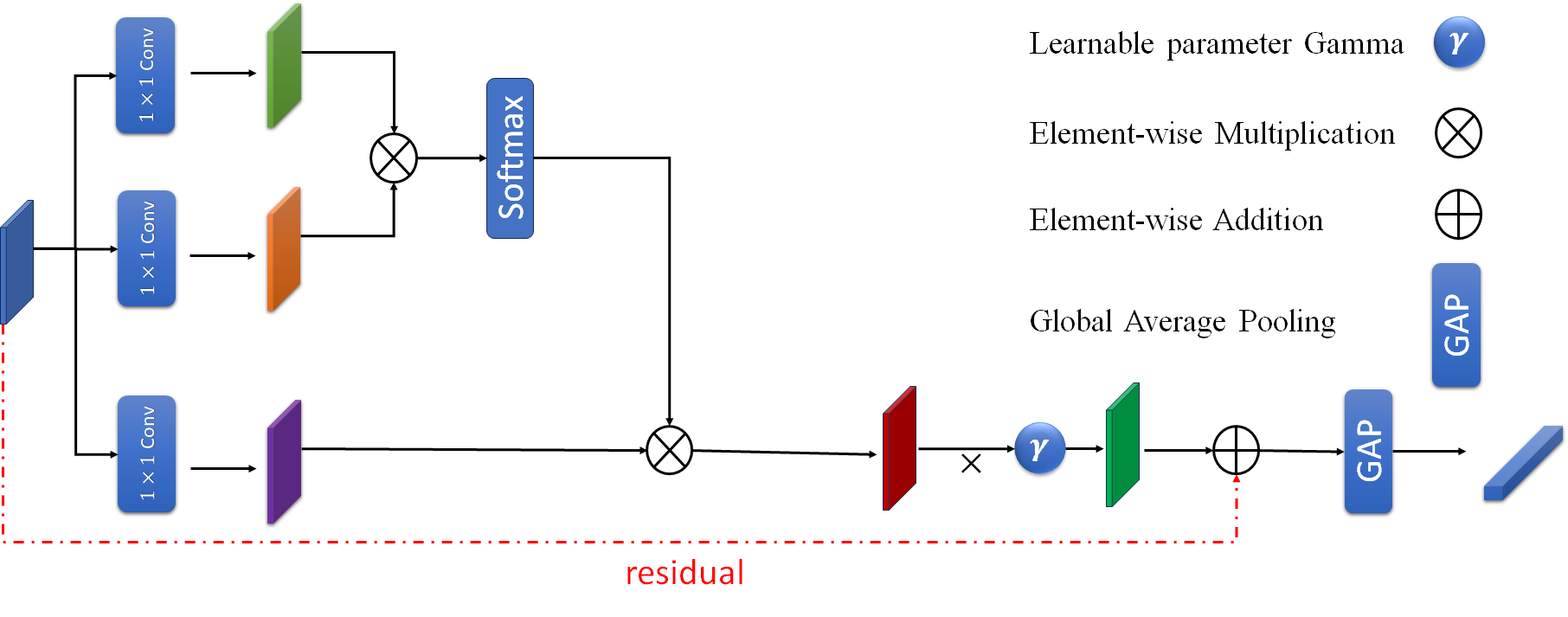}
	\caption{SA module}
	\label{SA module}
\end{figure*}

\subsubsection{\normalfont Attention module}
After extracting the feature vectors at each stage, we utilize self-attention and global average pooling. The representation of the attention module is illustrated in Figure~\ref{SA module} Considering the extracted feature vectors, we apply a $1 \times 1$ convolution on $f$, resulting in convolutional vectors $k$, $g$ and $h$. This operation is performed to reduce the number of channels (Equation \ref{eq_3}):
\begin{equation}
k',v',q' = Conv_{1\times1}(f^{Conv-p}), \; p\in [1-5]
\label{eq_3}
\end{equation}
After obtaining the convolutional vectors $q'(f^{Conv-p})$ and $k'(f^{Conv-p})$, we apply the softmax function (Equation \ref{eq_4}):
\begin{equation}
\begin{split}
\beta_{i,j} = \frac{exp(S_{i,j})}{\sum_{i=1}^{N}exp(S_{i,j})}, \; \;\;\; where \; \;\;\; S_{i,j} = \\ \\ q'(f_i^{Conv-p})^{T}k'(f_j^{Conv-p})
\end{split}
\label{eq_4}
\end{equation}
The attention mechanism computes weights $\beta$, that determine the relative importance of each pixel in the feature map. These weights are calculated across the entire spatial extent, allowing the attention module to capture long-range dependencies beyond a local neighborhood, unlike traditional convolutions. The scaling factor, typically denoted as $\gamma$, is a learnable parameter in the network that is multiplied with the input feature $f^{Conv-p}$ before the addition (Equation \ref{eq_6}):
\begin{equation}
y^{Conv-p} = \gamma \times \beta_i + f^{Conv-p}
\label{eq_6}
\end{equation}
After obtaining the final output, we take a global average pooling. The output is represented for each query and support sample as follows:
\begin{equation}
    {f'}_{i{j}}^{s} = \{ {f'}_{i{j}}^{Conv1-s},{f'}_{i{j}}^{Conv2-s},...,{f'}_{i{j}}^{Conv5-s} \}
	\label{eq_7}
\end{equation}
\begin{equation}
	{f'}_{i}^{q} = \{ {f'}_{i}^{Conv1-q},{f'}_{i}^{Conv2-q},...,{f'}_{i}^{Conv5-q} \}
	\label{eq_8}
\end{equation}

\subsubsection{\normalfont Distance metric}
Once we have obtained the final output from the previous stage, we take the average of the vectors from all the support samples belonging to the same class to obtain the prototypes for each class from the support set as Equation \ref{eq_9}.
\begin{equation}
	{c'}_{i}^{Conv-p} = \frac{1}{|K|}\sum_{j=1}^{K}{f'}_{i{j}}^{Conv-p-s}, \; where \; p \in [1-5]
	\label{eq_9}
\end{equation}
We calculate the Euclidean distance between the feature map of each query sample \( f_i^{'Conv-p} \) and its corresponding prototype map \( c_j^{'Conv-p} \), considering 5 feature maps per sample, as described in Equation \ref{eq_10}.
\begin{equation}
d_{i, j}^{\text {Conv-p }}=\operatorname{Euclidean}\left(f_i^{' C o n v-p}, c_j{ }^{'C o n v-p}\right)
\label{eq_10}
\end{equation}

\subsubsection{\normalfont Learnable weights}
Considering the emphasis of shallow network layers on global features and deep layers on abstract features, we opted to assign weights to the distances computed in the five feature spaces for each query sample-support set representative pair. These weights are trainable within the network, and their initial assignment significantly influences the model's performance, which will be elaborated on in the Experiment section. The final distance is obtained by aggregating the weighted distances from these five feature spaces as shown in Equation \ref{eq_11}:
\begin{equation}
	d_{i,j} = \sum_{r=1}^{5}w_{i,j}^{Conv-p}\times d_{i,j}^{Conv-p}, \; p\in [1-5]
	\label{eq_11}
\end{equation}
Afterward, we apply softmax as shown in Equation \ref{eq_13}:
\begin{equation}
	p(y=j|x_{i}^{q})= \frac{exp(-d_{i,j})}{\sum_{l=1}^{n_s}exp(-d_{i,l})}
	\label{eq_13}
\end{equation}
The learning process involves minimizing the negative log-probability \( J = -\log p(y=j|x_i^q) \) of the true class \( j \) using the Adam optimizer. Training episodes are formed by randomly selecting a subset of classes from the training set. Within each selected class, a subset of examples is chosen to form the support set, while another subset from the remaining examples is used as query points.

\section{Experimental Results}
\subsection{Datasets}

The proposed method was evaluated on three widely used datasets commonly employed for few-shot learning tasks: MiniImageNet \cite{bib34} and FC100 \cite{bib35}, as well as eight datasets from the CD-FSL benchmark, including TierImageNet \cite{bib80}, CUB \cite{wah2011caltech}, ChestX-ray \cite{bib81}, ISIC \cite{bib82}, Flower102 \cite{bib84}, EuroSAT (Euro) \cite{bib83}, CropDisease (CropD) \cite{bib85}, and histopathological image dataset \cite{bolhasani2020histopathological}.

\textbf{MiniImageNet dataset.} The MiniImageNet is a subset of the ImageNet dataset designed for training and evaluating machine learning models. This dataset contains 100 classes with 600 images per class. The images are randomly divided into three splits: 64 classes for training, 16 classes for validation, and 20 classes for testing. This data partitioning allows researchers to evaluate how effectively models can generalize to new and unseen classes after being trained on the provided set of classes.

\textbf{FC100 dataset.} The FC100 is another dataset similar in structure to MiniImageNet. It contains 100 classes with 600 images per class. However, the class splits are handled differently - the 100 classes are randomly divided into 60 training classes, 20 validation classes, and 20 test classes. This ensures that the training, validation, and test sets are entirely disjoint, which can provide a more realistic evaluation of a model's ability to learn general visual representations.

\subsection{Experimental Setting}
We performed standard preprocessing and model configuration steps for our datasets. All input images were resized to \( 84 \times 84 \) pixels and normalized using a standard normalization technique to improve model convergence. We employed a pre-trained ResNet-18 as the backbone feature extractor and utilized the Adam optimizer for model training. Learning rates and other specific settings are summarized in Table~\ref{tab:experimental_settings}.To ensure the reliability of our results, the code was run three times, and the result was reported. All experiments were conducted on an NVIDIA RTX 4090 GPU system.
\begin{table}[h]
\centering
\caption{Experimental Settings}
\label{tab:experimental_settings}
\resizebox{0.8\linewidth}{!}{
\renewcommand{\arraystretch}{1.3}
\begin{tabularx}{\linewidth}{|>{\centering\arraybackslash}X|>{\centering\arraybackslash}X|}
\hline
\textbf{Parameter} & \textbf{Configuration} \\ \hline
Input Image Resolution & \( 84 \times 84 \) pixels \\ \hline
Model Architecture & Pre-trained ResNet-18 \\ \hline
Feature Extraction & Feature maps from 5 stages \\ \hline
Optimizer & Adam \\ \hline
Learning Rate (MinImageNet) & \( 1 \times 10^{-4} \) \\ \hline
Learning Rate (FC100) & \( 2 \times 10^{-5} \) \\ \hline
Hardware & NVIDIA RTX 4090 GPU \\ \hline
\end{tabularx}
}
\end{table}

\subsection{Evaluation metric}
To evaluate the performance of our models, we employed the following scenario: For the training phase, we sampled 30 random classes and 5 examples per class from the training set. We then trained the model on these samples. For evaluation, we tested the trained model on a 5-way 5-shot task by selecting 5 random classes and using 5 examples per class. This process was repeated over multiple episodes to compute the overall 5-way 5-shot accuracy. Additionally, we evaluated the models on a 5-way 1-shot task. In this setting, we followed a similar approach but used only 1 example per class during the evaluation phase. Our primary evaluation metric for both tasks was accuracy.

\begin{table*}[h]
	\centering
	\caption{Evaluation on MiniImageNet in 5-way.}
	\label{tb:em}
	\renewcommand{\arraystretch}{1.3} % Adjust the space between rows
	\begin{tabular}{lccccc}
		\hline
		Method  & Year & Backbone & 1-shot & 5-shot \\
		\hline
		AdaResNet \cite{bib37}  & 2018 & ResNet12 & 56.88 & 71.94 \\
		TADAM \cite{bib38} & 2018 & ResNet12 & 58.50 & 76.70 \\
		LEO \cite{bib46} & 2018 & WRN-28-10 & 61.76 & 77.59 \\
		MetaOptNet \cite{bib39} & 2019 & ResNet12 & 62.64 & 78.63 \\
		CC+rot \cite{bib47} & 2019 & WRN-28-10 & 62.93 & 79.87 \\
		Neg-Cosine \cite{bib49} & 2020 & ResNet18 & 62.33 & 80.94 \\
		MixtFSL \cite{bib50} & 2020 & ResNet18 & 60.11 & 77.76 \\
		FEAT \cite{bib48} & 2020 & WRN-28-10 & 65.10 & 81.11 \\
		Neg-Margin \cite{bib40} & 2020 & ResNet12 & 63.85 & 81.57 \\
		Distill \cite{bib45} & 2020 & ResNet12 & 64.82 & 82.14 \\
		FRN \cite{bib43} & 2021 & ResNet12 & 66.45 & 82.83 \\
		MixtFSL \cite{bib41} & 2021 & ResNet12 & 63.98 & 82.04 \\
		Meta-Baseline \cite{bib42} & 2021 & ResNet12 & 63.17 & 79.26 \\
		MergeNet-Concat \cite{bib52} & 2021 & ResNet18 & 65.05 & 77.76 \\
		ViTFSL-Baseline \cite{bib67} & 2022 & Transformer & 63.51 & 80.30 \\
		QSFormer \cite{bib44} & 2023 & ResNet12 & 65.24 & 79.96 \\
		Meta-Baseline + DiffKendall \cite{bib51} & 2023 & ResNet12 & 65.56 & 80.79 \\
		Auto-MS \cite{bib70} & 2024 & HCE-64F & 53.33 & 69.64 \\
		Our model & 2024 & ResNet18 & \textbf{66.57} & \textbf{84.42} \\
		\hline
	\end{tabular}
\end{table*}

\begin{table}[h]
	\centering
	\caption{Evaluation on FC100 in 5-way.}
	\label{tb:ef}
	\renewcommand{\arraystretch}{1.3} % Adjust the space between rows
	\setlength{\tabcolsep}{2.pt} % Adjust column separation 
	\begin{tabular}{lccccc}
		\hline
		Method & Year & Backbone & 1-shot & 5-shot \\
		\hline
		P-Net \cite{bib30} & 2017 & ResNet12 & 37.80 & 53.30 \\
		TADAM \cite{bib54} & 2018 & ResNet12 & 40.10 & 56.10 \\
		Cosine Classifier \cite{bib53} & 2019 & ResNet12 & 38.47 & 57.67 \\
		SimpleShot \cite{bib55} & 2019 & ResNet10 & 40.13 & 53.63 \\
		Metaopt Net \cite{bib56} & 2019 & ResNet12 & 41.10 & 55.50 \\
		DC \cite{bib57} & 2019 & ResNet12 & 42.04 & 57.63 \\
		RFS \cite{bib60} & 2020 & ResNet12 & 44.60 & 60.90 \\
		MDM-Net \cite{bib58} & 2022 & ResNet12 & 43.62 & 57.41 \\
		SSFormers \cite{bib59} & 2023 & ResNet12 & 43.72 & 58.92 \\
		LSFSL \cite{bib65} & 2023 & ResNet12 & 43.60 & 60.12 \\
		Barlow Twins + DSA \cite{bib66} & 2024 & ViT-B & 41.42 & 55.47 \\
		Our model & 2024 & ResNet18 & \textbf{44.78} & \textbf{66.27} \\
		\hline
	\end{tabular}
\end{table}

\begin{table}[h]
	\centering
	\caption{The comparison of our model's performance against state-of-the-art methods on CUB and TieredImageNet.}
	\label{tab:cub_tiered}
	\renewcommand{\arraystretch}{1.3} % Adjust the space between rows
	\resizebox{0.5\textwidth}{!}{
		\begin{tabular}{>{\centering\arraybackslash}m{3cm}|cc|cc}
			\hline
			Method  & \multicolumn{2}{c|}{TieredImageNet} & \multicolumn{2}{c}{CUB} \\
			\cline{2-5}
			& 1-shot & 5-shot & 1-shot & 5-shot \\
			\hline
			MAML \cite{bib20}  & 51.61 & 65.76 & 40.51 & 53.09 \\
			ANIL \cite{bib72} & 52.82 & 66.52 & 41.12 & 55.82 \\
			BOIL \cite{bib76} & 53.23 & 69.37 & 44.20 & 60.92 \\
			Sparse-MAML \cite{bib77} & 53.47 & 68.83 & 41.37 & 60.58 \\
			Sparse-ReLU-MAML \cite{bib77}  & 53.77 & 68.12 & 42.89 & 57.33 \\
			Sparse-MAML+ \cite{bib77} & 53.91 & 69.92 & 43.43 & 62.02 \\
			GAP \cite{bib78} & 58.56 & 72.82 & 44.74 & 64.88 \\
			Our model  & \textbf{67.40} & \textbf{85.67} & \textbf{52.95} & \textbf{71.59} \\
			\hline
		\end{tabular}
	}
\end{table}

\begin{table}[h]
	\centering
	\caption{The comparison of our model's performance against state-of-the-art methods on the test domains of selected datasets in the 5-way 5-shot task.}
	\label{tb:selected_methods_performance}
	\renewcommand{\arraystretch}{1.3} % Adjust the space between rows
	\resizebox{0.5\textwidth}{!}{ % Resize the table to fit the text width
		\begin{tabular}{lccccc}
			\hline
			Method       & ChessX & CropD & Euro & Flower & ISIC \\
			\hline
			BL++ \cite{bib53}     & 25.49 & 48.22 & 75.79 & 63.16 & 40.73 \\
			ANIL \cite{bib72}      & 24.41 & 48.69 & 63.96 & 61.27 & 37.57 \\
			ANIL+MLP \cite{bib73}  & 25.02 & 58.10 & 75.73 & 66.45 & 39.19 \\
			MTL \cite{bib28}      & 24.15 & 33.27 & 54.27 & 58.18 & 35.56 \\
			MTL+MLP \cite{bib73}  & 25.19 & 51.23 & 65.19 & 51.34 & 34.74 \\
			PN \cite{bib30}       & 24.05 & 55.59 & 70.96 & 63.56 & 32.95 \\
			DN4 \cite{bib74}      & 27.34 & 53.62 & 75.01 & 75.01 & 40.15 \\
			CAN \cite{bib75}      &  \textbf{27.46} & 67.26 & 78.22 & 79.81 & 42.73 \\
			Our model  & 26.33 & \textbf{83.48} & \textbf {78.63} & \textbf {85.13} &  \textbf {43.02} \\
			\hline
		\end{tabular}
	}
\end{table}

\begin{table}[h]
	\centering
	\caption{The comparison of our model's performance on histopathological image dataset \cite{bolhasani2020histopathological} in 3-way 1-shot and 3-way 5-shot tasks.}
	\label{tb:histopathological}
	\renewcommand{\arraystretch}{1.3} % Adjust the space between rows
	%\resizebox{0.5\textwidth}{!}{ % Resize the table to fit the text width
		\begin{tabular}{lcc}
			\hline
			Method       & 1-shot & 5-shot  \\
			\hline
			Baseline     & 38.42 & 42.65  \\

			Proposed method  & \textbf {39.17} & \textbf{43.94} \\
			
			\hline
		\end{tabular}
	%}
\end{table}

\begin{table*}[h]
	\centering
	\caption{The influence of each component to the model’s performance}
	\label{tb:ec}
	\renewcommand{\arraystretch}{1.3} % Adjust the space between rows
	\resizebox{0.8\textwidth}{!}{
		\begin{tabular}{|c|c|c|c|cc|cc|c|c|}
			\hline
			\multirow{2}{*}{Baseline} &
			\multirow{2}{*}{Multiscale} &
			\multirow{2}{*}{\begin{tabular}[c]{@{}c@{}}Learnable\\ Weight\end{tabular}} &
			\multirow{2}{*}{\begin{tabular}[c]{@{}c@{}}Self-\\ attention\end{tabular}} &
			\multicolumn{2}{c|}{MiniImageNet} &
			\multicolumn{2}{c|}{FC100} & \multicolumn{2}{c|}{Additional Metrics} \\ \cline{5-10} 
			&  &  &  & \multicolumn{1}{c|}{1 shot} & \multicolumn{1}{c|}{5 shot} & \multicolumn{1}{c|}{1 shot} & \multicolumn{1}{c|}{5 shot} & \multicolumn{1}{c|}{timeInference} & \multicolumn{1}{c|}{numParameter} \\ \hline
			\checkmark & X           & X           & X           & 62.83 & 82.38 & 39.47 & 63.44 & 0.145s & 11M \\
			\checkmark & \checkmark  & X           & X           & 63.65 & 83.23 & 41.97 & 64.80 & 0.377s & 11M \\
			\checkmark & \checkmark  & \checkmark  & X           & 64.73 & 83.92 & 43.70 & 65.76 & 0.380s & 11M \\
			\checkmark & \checkmark  & \checkmark  & \checkmark  & \textbf{66.57} & \textbf{84.42} & \textbf{44.78} & \textbf{66.27} & \textbf{0.438s} & 12M \\
			\hline
		\end{tabular}
	}
\end{table*}

\begin{table*}[h]
	\centering
	\caption{The influence of different weights at each stage on the model's performance on MiniImageNet}
	\label{tb:dw}
	\renewcommand{\arraystretch}{1.5} % Adjust the space between rows
	\resizebox{\textwidth}{!}{
		
		\begin{tabular}{|c|ccccc|cc|cc|}
			\hline
			\multirow{2}{*}{} &
			\multicolumn{5}{c|}{Weights} &
			\multicolumn{2}{c|}{1 shot} &
			\multicolumn{2}{c|}{5 shot} \\ \cline{2-10} 
			&
			\multicolumn{1}{c|}{w1} &
			\multicolumn{1}{c|}{w2} &
			\multicolumn{1}{c|}{w3} &
			\multicolumn{1}{c|}{w4} &
			w5 &
			\multicolumn{1}{c|}{\begin{tabular}[c]{@{}c@{}}accuracy without\\ learnable weights\end{tabular}} &
			\begin{tabular}[c]{@{}c@{}}accuracy with\\ learnable weights\end{tabular} &
			\multicolumn{1}{c|}{\begin{tabular}[c]{@{}c@{}}accuracy without\\ learnable weights\end{tabular}} &
			\begin{tabular}[c]{@{}c@{}}accuracy with\\ learnable weights\end{tabular} \\ \hline
			\multirow{3}{*}{\begin{tabular}[c]{@{}c@{}}Weight\\ Initialization\end{tabular}} &
			\multicolumn{1}{c|}{1} &
			\multicolumn{1}{c|}{1} &
			\multicolumn{1}{c|}{1} &
			\multicolumn{1}{c|}{1} &
			1 &
			\multicolumn{1}{c|}{63.80\%} &
			64.54\% &
			\multicolumn{1}{c|}{82.24\%} &
			82.99\% \\ \cline{2-10} 
			&
			\multicolumn{1}{c|}{1} &
			\multicolumn{1}{c|}{1.1} &
			\multicolumn{1}{c|}{1.2} &
			\multicolumn{1}{c|}{1.3} &
			1.4 &
			\multicolumn{1}{c|}{63.34\%} &
			\textbf{64.73\%} &
			\multicolumn{1}{c|}{83.02\%} &
			\textbf{83.92\%} \\ \cline{2-10} 
			&
			\multicolumn{1}{c|}{1} &
			\multicolumn{1}{c|}{1.2} &
			\multicolumn{1}{c|}{1.4} &
			\multicolumn{1}{c|}{1.6} &
			1.8 &
			\multicolumn{1}{c|}{62.13\%} &
			64.6\% &
			\multicolumn{1}{c|}{83.68\%} &
			83.70\% \\ \cline{2-10} 
                &
			\multicolumn{1}{c|}{1} &
			\multicolumn{1}{c|}{0.9} &
			\multicolumn{1}{c|}{0.8} &
			\multicolumn{1}{c|}{0.7} &
			0.6 &
			\multicolumn{1}{c|}{63.2\%} &
			{62.36\%} &
			\multicolumn{1}{c|}{80.49\%} &
			{80.29\%} \\ \cline{2-10} 
               &
               \multicolumn{1}{c|}{1} &
			\multicolumn{1}{c|}{0.8} &
			\multicolumn{1}{c|}{0.6} &
			\multicolumn{1}{c|}{0.4} &
			0.2 &
			\multicolumn{1}{c|}{62.36\%} &
			{61.24\%} &
			\multicolumn{1}{c|}{78.96\%} &
			{78.84\%} \\ \hline 
		\end{tabular}
	}
\end{table*}

\subsection{Comparison with state of the arts}
Based on the results presented in Table \ref{tb:em} and Table \ref{tb:ef}, our proposed model demonstrates strong few-shot learning performance compared to the state-of-the-art approaches. As shown in Table \ref{tb:em}, on the MiniImageNet dataset, our model achieves an accuracy of 66.57 in the 1-shot setting and 84.42 in the 5-shot setting. This indicates that our model is able to effectively leverage the limited training data and rapidly adapt to new tasks, showcasing its superior few-shot learning capabilities.Similarly, on the more challenging FC100 dataset, as shown in Table \ref{tb:ef}, our model outperforms the existing state-of-the-art methods by a significant margin, obtaining an accuracy of 44.78 in the 1-shot setting and 66.27 in the 5-shot setting. The consistent improvements observed across both MiniImageNet and FC100 datasets highlight the effectiveness of the design choices and techniques employed in our model. These choices and techniques allow it to learn more robust and transferable representations for few-shot learning scenarios.These results position our model as a highly competitive approach in the field of few-shot learning.

\subsection{Cross domain}
To further evaluate the generalization capabilities of our proposed model, we conducted a cross-domain evaluation using the MiniImageNet dataset for training and eight medical datasets as testing domains. These datasets span various medical imaging tasks, providing a comprehensive benchmark for assessing how well models trained on general-purpose datasets can generalize to new and specialized domains. 

Table~\ref{tb:selected_methods_performance} compares our model’s performance with state-of-the-art methods, including BL++, ANIL, CAN, DN4, and MTL+MLP, on challenging 5-way 5-shot tasks. Table~\ref{tab:cub_tiered} further shows performance comparisons on widely used datasets such as CUB and TieredImageNet. Our model demonstrated strong generalization capabilities, outperforming or matching baseline methods in multiple cases. 

Additionally, we evaluated our proposed method on a histopathological image dataset \cite{bolhasani2020histopathological}, which contains three grades and thus represents a 3-way classification problem. Table~\ref{tb:histopathological} summarizes the results of this comparison for both 1-shot and 5-shot tasks. 

The results indicate that our method successfully adapted to diverse testing domains, showing good performance in tasks such as skin lesion classification, flower identification, and crop disease detection. These findings underscore the effectiveness of our approach in handling domain shifts and limited labeled data.

\subsection{Ablation study}
To better understand the contributions of different components in our model, we conducted an ablation study and reported the results. Table \ref{tb:ec} illustrates the influence of individual components on overall model performance. The Baseline configuration, serving as the foundation of our model, achieves notable results with 62.83\% accuracy for the 1-shot task and 82.38\% for the 5-shot task on the MiniImageNet dataset. On the FC100 dataset, the Baseline configuration results in 39.47\% 1-shot and 63.44\% 5-shot accuracy.

Introducing the Multiscale module, which enhances feature extraction by capturing information at multiple scales, leads to improvements in both 1-shot and 5-shot performance across both datasets. Adding the Learnable Weight component, which adapts weights for different feature channels, further boosts accuracy, demonstrating its effectiveness.

The final inclusion of the Self-attention module results in the best overall performance, with the model achieving 66.57\% accuracy for the 1-shot task and 84.42\% for the 5-shot task on MiniImageNet. On FC100, the performance reaches 44.78\% for 1-shot and 66.27\% for 5-shot tasks. These results underscore the significant contribution of the self-attention mechanism to enhancing the few-shot learning capabilities of our approach.

In addition to the improved accuracy, we evaluated the efficiency of the proposed model by comparing the number of parameters and inference time with the baseline. As shown in Table \ref{tb:ec}, the final model includes one million additional parameters compared to the baseline. Despite this slight increase, the accuracy shows significant improvements across all tasks. Furthermore, the inference time of the final model is approximately three times that of the baseline, which remains a reasonable trade-off considering the substantial accuracy gains. These results highlight the effectiveness of our approach in achieving enhanced performance while maintaining computational efficiency.

As shown in Table \ref{tb:dw}, two important observations can be made regarding model performance on the MiniImageNet dataset:

1) Advantage of Learnable Weights: The results indicate a clear benefit to using learnable weights. In both the 5-way 1-shot and 5-way 5-shot tasks, models with learnable weights achieved higher accuracies compared to those with fixed weights. For example, in the 5-way 1-shot task, accuracy increased from 63.80\% to 64.54\% when weights were learnable. Similarly, in the 5-way 5-shot task, accuracy improved from 82.24\% to 82.99\% with learnable weights. These results demonstrate that incorporating learnable weights significantly enhances model performance.

2) The choice of weight initialization plays a crucial role in model performance. To investigate this, we started with equal initial weights of 1,1,1,1,1 for all feature vectors. From this starting point, we incrementally increased the weight of each feature vector by 0.1 relative to the previous stage. Among all these configurations, the best results were observed with initial weights of 1,1.1,1.2,1.3,1.4. This process underscores the importance of proper weight initialization, demonstrating that small and systematic adjustments to initial weights can lead to significant improvements in model performance. Furthermore, as shown in the final two rows of Table \ref{tb:dw}, we also experimented with assigning lower initial weights to deep layers compared to shallow layers. This approach resulted in lower accuracy than configurations where deep layers were assigned higher weights. These findings suggest that deep layers play a more significant and impactful role in classification tasks compared to shallow layers.

In summary, Table \ref{tb:dw}, highlights that both the use of learnable weights and careful selection of weight initialization are key factors in improving model performance.

\begin{table}[h!]
	\centering
	\caption{The influence of different gamma at each stage on the model's performance on MiniImageNet}
	\label{tb:dg}
	\renewcommand{\arraystretch}{1.3} % Adjust the space between rows
	\resizebox{\columnwidth}{!}{
		\begin{tabular}{|c|c|c|c|}
			\hline
			Gamma 1 & Gamma 2 & 1 shot           & 5 shot           \\ \hline
			0.2     & 0.2     & \textbf{66.57\%} & \textbf{84.42\%} \\ \hline
			0.3     & 0.3     & 65.09\%          & 84.12\%          \\ \hline
			0.4     & 0.4     & 65.15\%          & 84.16\%          \\ \hline
		\end{tabular}
	}
\end{table}

As shown in Table \ref{tb:dg}, the impact of different \(\gamma\) values on model performance in the self-attention module is evident for both 1-shot and 5-shot tasks. Specifically, setting \(\gamma\) values to 0.2 for both support and query images resulted in the highest accuracy. This indicates that the choice of \(\gamma\) significantly influences model performance, with the 0.2 setting yielding the best results compared to other tested values.

\subsection{Analysis}
In this section, we present a detailed evaluation of the model's predictions on the MiniImageNet dataset, incorporating both quantitative and qualitative analyses. Visual examples are used to highlight correctly classified and misclassified samples, offering insights into the model's strengths and limitations in decision-making. Additionally, these visualizations help uncover the underlying factors contributing to correct classifications and prediction errors.

\begin{figure}[ht]
    \centering
    \includegraphics[width=1\linewidth]{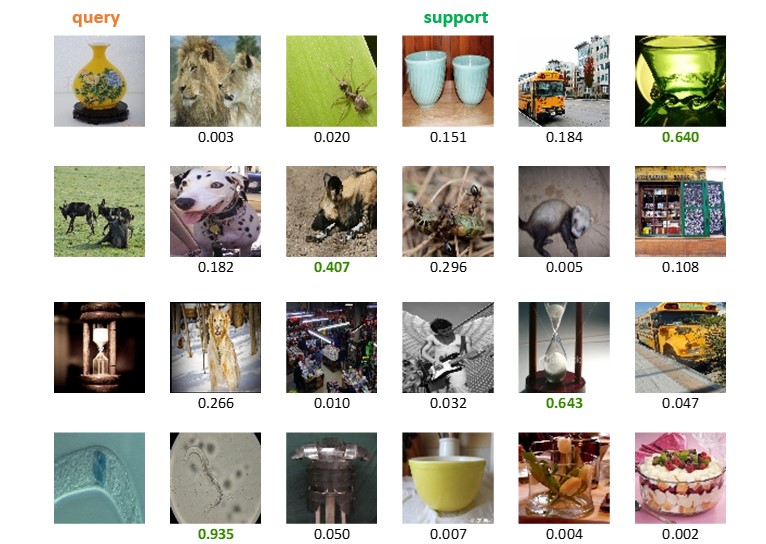}
    \caption{Examples of correctly classified samples on the MiniImageNet dataset.}
    \label{fig:correct_samples}
\end{figure}

Figure~\ref{fig:correct_samples} displays correctly classified samples. For example, in the \textbf{first row}, despite similarities between the query image of a vase and support images such as cups, the model successfully classifies the vase with a confidence of \textbf{64\%}. This demonstrates the model's capacity to distinguish fine-grained details despite visual similarities. Additionally, the green-highlighted prediction scores underscore the model’s robust generalization across challenging query-support pairs.

\begin{figure}[ht]
    \centering
    \includegraphics[width=1\linewidth]{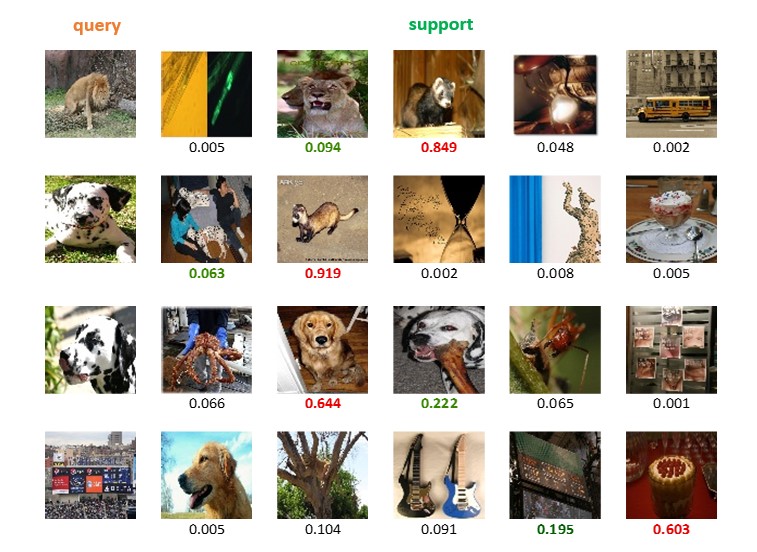}
    \caption{Examples of misclassified samples on the MiniImageNet dataset.}
    \label{fig:misclassified_samples}
\end{figure}

Conversely, Figure~\ref{fig:misclassified_samples} illustrates instances where the model makes incorrect predictions. Upon examination, these misclassifications reveal key challenges in visual reasoning:

\begin{itemize}
    \item In the \textbf{third row}, a black-and-white spotted dog in the query image is misclassified as a different breed of dog, likely due to their visual resemblance.
    \item Similarly, in other cases, objects with overlapping features or ambiguous contexts appear to confuse the model, leading to reasonable but incorrect predictions.
\end{itemize}

These observations demonstrate that while the model effectively handles complex cases in many instances, future improvements could further enhance its ability to distinguish visually similar categories and reduce context-driven errors. Addressing these issues is crucial for achieving more accurate classification performance under challenging conditions.

\begin{figure*}[h]
	\centering
	\includegraphics[width=1\textwidth]{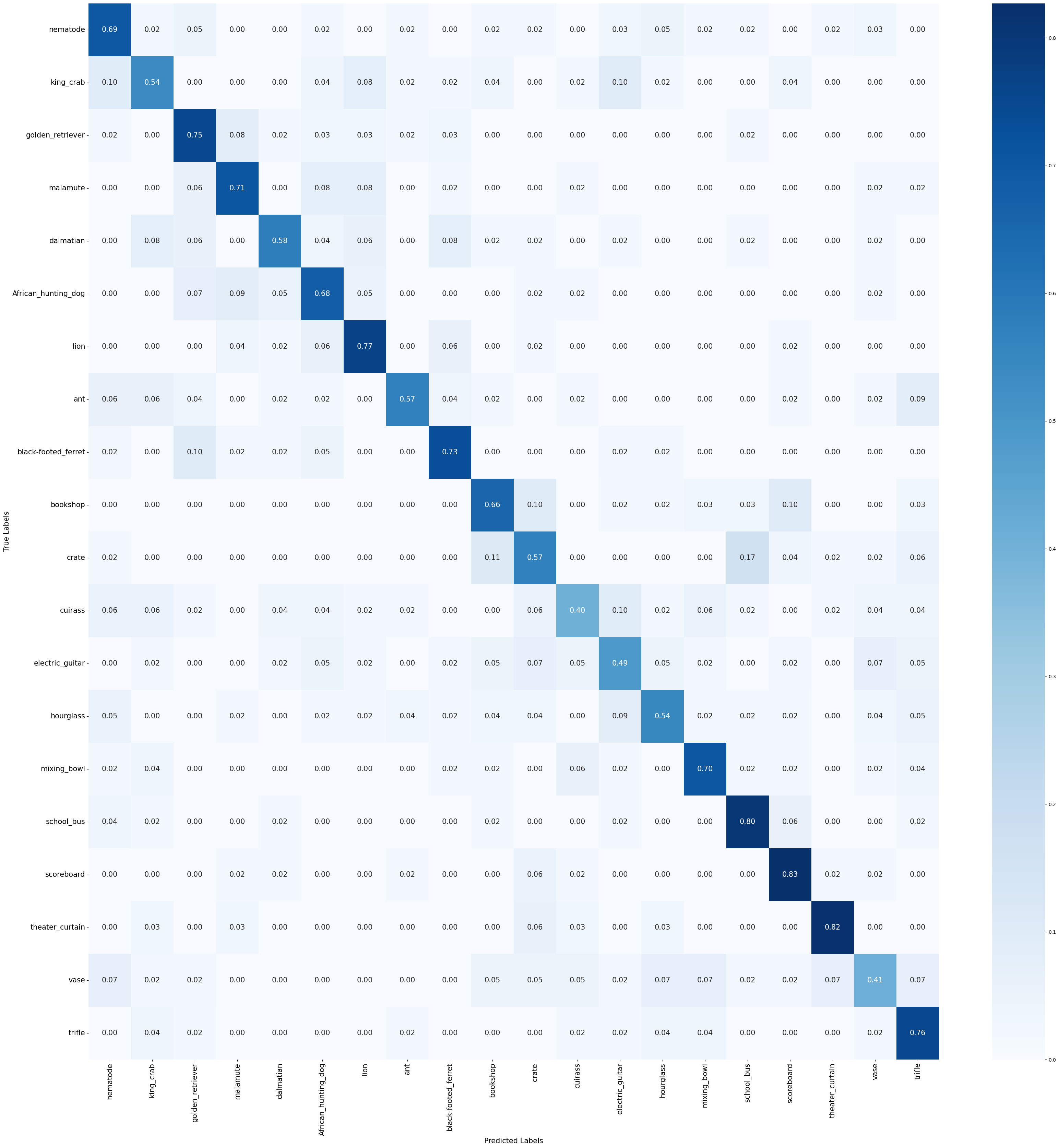}
	\caption{Confusion matrix for the model's predictions on the MiniImageNet dataset for the 5-way 1-shot task.}
	\label{confusion_matrix}
\end{figure*}

Figure~\ref{confusion_matrix} illustrates the confusion matrix for our model on the MiniImageNet dataset, highlighting the class-wise prediction performance and potential misclassification patterns. The results reveal that the model generally distinguishes classes well, but certain visually similar categories present challenges. 

For instance, the golden retriever and dalmatian classes show occasional confusion due to shared visual characteristics such as similar body structures and fur patterns. Similarly, the electric guitar class is sometimes misclassified as curiass, likely due to overlapping visual features such as elongated shapes and complex backgrounds. These cases reflect reasonable misclassifications given the inherent visual similarities between query and support images. Despite these challenges, the overall distribution of correct classifications demonstrates the robustness of the model across diverse categories.

\section{Conclusion}
This paper presents an innovative strategy to enhance few-shot classification by integrating a self-attention network and embedding learnable weights at each stage, leading to improved performance and significant outcomes. By employing feature vector extraction and weight transfer across stages, our approach elevates multi-scale feature representation, resulting in enhanced overall model performance. The incorporation of self-attention mechanisms effectively refines features at each stage, yielding more robust representations. Extensive evaluations on the MiniImageNet and FC100 datasets demonstrate the efficacy of our method compared to current state-of-the-art approaches. To further validate our model's generalization capabilities, we conducted experiments across eight cross-domain datasets.Future work will focus on exploring the theoretical rationale behind weight initialization for each stage, which is crucial for optimizing model performance. We also propose a two-phase training approach that eliminates less relevant support images during the initial training phase, allowing the model to concentrate on those most closely related to the query image, ultimately enhancing performance.Furthermore, our method can be adapted and extended to other few-shot learning tasks beyond image classification, such as few-shot object detection and segmentation. To handle larger and more complex datasets, modifications may include refining attention mechanisms and optimizing weight initialization strategies to accommodate increased variability and complexity in the data.
% Non-BibTeX users please use

%\pagebreak
\bibliographystyle{IEEEtran}
\bibliography{ref1.bib}

% Generated by IEEEtran.bst, version: 1.14 (2015/08/26)
\begin{thebibliography}{10}
\providecommand{\url}[1]{#1}
\csname url@samestyle\endcsname
\providecommand{\newblock}{\relax}
\providecommand{\bibinfo}[2]{#2}
\providecommand{\BIBentrySTDinterwordspacing}{\spaceskip=0pt\relax}
\providecommand{\BIBentryALTinterwordstretchfactor}{4}
\providecommand{\BIBentryALTinterwordspacing}{\spaceskip=\fontdimen2\font plus
\BIBentryALTinterwordstretchfactor\fontdimen3\font minus
  \fontdimen4\font\relax}
\providecommand{\BIBforeignlanguage}[2]{{%
\expandafter\ifx\csname l@#1\endcsname\relax
\typeout{** WARNING: IEEEtran.bst: No hyphenation pattern has been}%
\typeout{** loaded for the language `#1'. Using the pattern for}%
\typeout{** the default language instead.}%
\else
\language=\csname l@#1\endcsname
\fi
#2}}
\providecommand{\BIBdecl}{\relax}
\BIBdecl

\bibitem{bib1}
A.~Saber, P.~Parhami, A.~Siahkarzadeh, and A.~Fateh, ``Efficient and accurate
  pneumonia detection using a novel multi-scale transformer approach,''
  \emph{arXiv preprint arXiv:2408.04290}, 2024.

\bibitem{fateh2024advancing}
A.~Fateh, R.~T. Birgani, M.~Fateh, and V.~Abolghasemi, ``Advancing multilingual
  handwritten numeral recognition with attention-driven transfer learning,''
  \emph{IEEE Access}, vol.~12, pp. 41\,381--41\,395, 2024.

\bibitem{bib2}
A.~Sharif~Razavian, H.~Azizpour, J.~Sullivan, and S.~Carlsson, ``Cnn features
  off-the-shelf: an astounding baseline for recognition,'' pp. 806--813, 2014.

\bibitem{tian2024survey}
S.~Tian, L.~Li, W.~Li, H.~Ran, X.~Ning, and P.~Tiwari, ``A survey on few-shot
  class-incremental learning,'' \emph{Neural Networks}, vol. 169, pp. 307--324,
  2024.

\bibitem{sun2024klsanet}
Z.~Sun, W.~Zheng, and P.~Guo, ``Klsanet: Key local semantic alignment network
  for few-shot image classification,'' \emph{Neural Networks}, p. 106456, 2024.

\bibitem{rezvani2024fusionlungnet}
S.~Rezvani, M.~Fateh, Y.~Jalali, and A.~Fateh, ``Fusionlungnet: Multi-scale
  fusion convolution with refinement network for lung ct image segmentation,''
  \emph{arXiv preprint arXiv:2410.15812}, 2024.

\bibitem{bib7}
Y.~Wang, Q.~Yao, J.~T. Kwok, and L.~M. Ni, ``Generalizing from a few examples:
  A survey on few-shot learning,'' \emph{ACM computing surveys (csur)},
  vol.~53, no.~3, pp. 1--34, 2020.

\bibitem{wang2023data}
Q.~Wang, F.~Meng, and T.~P. Breckon, ``Data augmentation with norm-ae and
  selective pseudo-labelling for unsupervised domain adaptation,'' \emph{Neural
  Networks}, vol. 161, pp. 614--625, 2023.

\bibitem{bib9}
J.~Zhou, Y.~Zheng, J.~Tang, J.~Li, and Z.~Yang, ``Flipda: Effective and robust
  data augmentation for few-shot learning,'' \emph{arXiv preprint
  arXiv:2108.06332}, 2021.

\bibitem{bib10}
D.~Xue, X.~Zhou, C.~Li, Y.~Yao, M.~M. Rahaman, J.~Zhang, H.~Chen, J.~Zhang,
  S.~Qi, and H.~Sun, ``An application of transfer learning and ensemble
  learning techniques for cervical histopathology image classification,''
  \emph{IEEE Access}, vol.~8, pp. 104\,603--104\,618, 2020.

\bibitem{bib11}
H.~Xia, H.~Zhao, and Z.~Ding, ``Adaptive adversarial network for source-free
  domain adaptation,'' pp. 9010--9019, 2021.

\bibitem{li2023novel}
J.~Li, F.~Wang, H.~Huang, F.~Qi, and J.~Pan, ``A novel semi-supervised meta
  learning method for subject-transfer brain--computer interface,''
  \emph{Neural Networks}, vol. 163, pp. 195--204, 2023.

\bibitem{fateh2024msdnet}
A.~Fateh, M.~R. Mohammadi, and M.~R.~J. Motlagh, ``Msdnet: Multi-scale decoder
  for few-shot semantic segmentation via transformer-guided prototyping,''
  \emph{arXiv preprint arXiv:2409.11316}, 2024.

\bibitem{yang2024meta}
Z.~Yang, J.~Xia, S.~Li, W.~Liu, S.~Zhi, S.~Zhang, L.~Liu, Y.~Fu, and
  D.~G{\"u}nd{\"u}z, ``Meta-learning based bind image super-resolution approach
  to different degradations,'' \emph{Neural Networks}, p. 106429, 2024.

\bibitem{bib14}
M.~Goldblum, L.~Fowl, and T.~Goldstein, ``Adversarially robust few-shot
  learning: A meta-learning approach,'' \emph{Advances in Neural Information
  Processing Systems}, vol.~33, pp. 17\,886--17\,895, 2020.

\bibitem{bib15}
A.~Parnami and M.~Lee, ``Learning from few examples: A summary of approaches to
  few-shot learning,'' \emph{arXiv preprint arXiv:2203.04291}, 2022.

\bibitem{bib17}
W.~Bian, Y.~Chen, X.~Ye, and Q.~Zhang, ``An optimization-based meta-learning
  model for mri reconstruction with diverse dataset,'' \emph{Journal of
  Imaging}, vol.~7, no.~11, p. 231, 2021.

\bibitem{bib18}
H.~Cho, Y.~Cho, J.~Yu, and J.~Kim, ``Camera distortion-aware 3d human pose
  estimation in video with optimization-based meta-learning,'' pp.
  11\,169--11\,178, 2021.

\bibitem{liu2024few}
Y.~Liu, D.~Shi, and H.~Lin, ``Few-shot learning with representative global
  prototype,'' \emph{Neural Networks}, vol. 180, p. 106600, 2024.

\bibitem{bib20}
P.~Li, G.~Zhao, and X.~Xu, ``Coarse-to-fine few-shot classification with deep
  metric learning,'' \emph{Information Sciences}, vol. 610, pp. 592--604, 2022.

\bibitem{bib21}
F.~Gao, L.~Cai, Z.~Yang, S.~Song, and C.~Wu, ``Multi-distance metric network
  for few-shot learning,'' \emph{International Journal of Machine Learning and
  Cybernetics}, vol.~13, no.~9, pp. 2495--2506, 2022.

\bibitem{bib22}
A.~Afrasiyabi, H.~Larochelle, J.-F. Lalonde, and C.~Gagn{\'e}, ``Matching
  feature sets for few-shot image classification,'' pp. 9014--9024, 2022.

\bibitem{bib23}
Q.~Cai, Y.~Pan, T.~Yao, C.~Yan, and T.~Mei, ``Memory matching networks for
  one-shot image recognition,'' pp. 4080--4088, 2018.

\bibitem{bib24}
O.~Vinyals, C.~Blundell, T.~Lillicrap, D.~Wierstra \emph{et~al.}, ``Matching
  networks for one shot learning,'' \emph{Advances in neural information
  processing systems}, vol.~29, 2016.

\bibitem{bib25}
T.~Munkhdalai and H.~Yu, ``Meta networks,'' pp. 2554--2563, 2017.

\bibitem{bib26}
M.~Garnelo, D.~Rosenbaum, C.~Maddison, T.~Ramalho, D.~Saxton, M.~Shanahan,
  Y.~W. Teh, D.~Rezende, and S.~A. Eslami, ``Conditional neural processes,''
  pp. 1704--1713, 2018.

\bibitem{bib27}
C.~Finn, P.~Abbeel, and S.~Levine, ``Model-agnostic meta-learning for fast
  adaptation of deep networks,'' pp. 1126--1135, 2017.

\bibitem{bib28}
Q.~Sun, Y.~Liu, T.-S. Chua, and B.~Schiele, ``Meta-transfer learning for
  few-shot learning,'' pp. 403--412, 2019.

\bibitem{bib29}
G.~Koch, R.~Zemel, R.~Salakhutdinov \emph{et~al.}, ``Siamese neural networks
  for one-shot image recognition,'' vol.~2, no.~1, 2015.

\bibitem{bib30}
J.~Snell, K.~Swersky, and R.~Zemel, ``Prototypical networks for few-shot
  learning,'' \emph{Advances in neural information processing systems},
  vol.~30, 2017.

\bibitem{bib31}
F.~Sung, Y.~Yang, L.~Zhang, T.~Xiang, P.~H. Torr, and T.~M. Hospedales,
  ``Learning to compare: Relation network for few-shot learning,'' pp.
  1199--1208, 2018.

\bibitem{bib32}
X.~Wang, X.~Wang, B.~Jiang, and B.~Luo, ``Few-shot learning meets transformer:
  Unified query-support transformers for few-shot classification,'' \emph{IEEE
  Transactions on Circuits and Systems for Video Technology}, 2023.

\bibitem{bib79}
F.~Hao, F.~He, L.~Liu, F.~Wu, D.~Tao, and J.~Cheng, ``Class-aware patch
  embedding adaptation for few-shot image classification,'' in
  \emph{Proceedings of the IEEE/CVF International Conference on Computer
  Vision}, 2023, pp. 18\,905--18\,915.

\bibitem{bib34}
O.~Vinyals, C.~Blundell, T.~Lillicrap, D.~Wierstra \emph{et~al.}, ``Matching
  networks for one shot learning,'' \emph{Advances in neural information
  processing systems}, vol.~29, 2016.

\bibitem{bib35}
B.~Oreshkin, P.~Rodr{\'\i}guez~L{\'o}pez, and A.~Lacoste, ``Tadam: Task
  dependent adaptive metric for improved few-shot learning,'' \emph{Advances in
  neural information processing systems}, vol.~31, 2018.

\bibitem{bib80}
M.~Ren, E.~Triantafillou, S.~Ravi, J.~Snell, K.~Swersky, J.~B. Tenenbaum,
  H.~Larochelle, and R.~S. Zemel, ``Meta-learning for semi-supervised few-shot
  classification,'' \emph{arXiv preprint arXiv:1803.00676}, 2018.

\bibitem{wah2011caltech}
C.~Wah, S.~Branson, P.~Welinder, P.~Perona, and S.~Belongie, ``The caltech-ucsd
  birds-200-2011 dataset,'' 2011.

\bibitem{bib81}
X.~Wang, Y.~Peng, L.~Lu, Z.~Lu, M.~Bagheri, and R.~M. Summers, ``Chestx-ray8:
  Hospital-scale chest x-ray database and benchmarks on weakly-supervised
  classification and localization of common thorax diseases,'' in
  \emph{Proceedings of the IEEE conference on computer vision and pattern
  recognition}, 2017, pp. 2097--2106.

\bibitem{bib82}
N.~Codella, V.~Rotemberg, P.~Tschandl, M.~E. Celebi, S.~Dusza, D.~Gutman,
  B.~Helba, A.~Kalloo, K.~Liopyris, M.~Marchetti \emph{et~al.}, ``Skin lesion
  analysis toward melanoma detection 2018: A challenge hosted by the
  international skin imaging collaboration (isic),'' \emph{arXiv preprint
  arXiv:1902.03368}, 2019.

\bibitem{bib84}
M.-E. Nilsback and A.~Zisserman, ``Automated flower classification over a large
  number of classes,'' in \emph{2008 Sixth Indian conference on computer
  vision, graphics \& image processing}.\hskip 1em plus 0.5em minus 0.4em\relax
  IEEE, 2008, pp. 722--729.

\bibitem{bib83}
P.~Helber, B.~Bischke, A.~Dengel, and D.~Borth, ``Eurosat: A novel dataset and
  deep learning benchmark for land use and land cover classification,''
  \emph{IEEE Journal of Selected Topics in Applied Earth Observations and
  Remote Sensing}, vol.~12, no.~7, pp. 2217--2226, 2019.

\bibitem{bib85}
S.~P. Mohanty, D.~P. Hughes, and M.~Salath{\'e}, ``Using deep learning for
  image-based plant disease detection,'' \emph{Frontiers in plant science},
  vol.~7, p. 1419, 2016.

\bibitem{bib37}
T.~Munkhdalai, X.~Yuan, S.~Mehri, and A.~Trischler, ``Rapid adaptation with
  conditionally shifted neurons,'' in \emph{International conference on machine
  learning}.\hskip 1em plus 0.5em minus 0.4em\relax PMLR, 2018, pp. 3664--3673.

\bibitem{bib38}
B.~Oreshkin, P.~Rodr{\'\i}guez~L{\'o}pez, and A.~Lacoste, ``Tadam: Task
  dependent adaptive metric for improved few-shot learning,'' \emph{Advances in
  neural information processing systems}, vol.~31, 2018.

\bibitem{bib46}
A.~A. Rusu, D.~Rao, J.~Sygnowski, O.~Vinyals, R.~Pascanu, S.~Osindero, and
  R.~Hadsell, ``Meta-learning with latent embedding optimization,'' \emph{arXiv
  preprint arXiv:1807.05960}, 2018.

\bibitem{bib39}
K.~Lee, S.~Maji, A.~Ravichandran, and S.~Soatto, ``Meta-learning with
  differentiable convex optimization,'' in \emph{Proceedings of the IEEE/CVF
  conference on computer vision and pattern recognition}, 2019, pp.
  10\,657--10\,665.

\bibitem{bib47}
S.~Gidaris, A.~Bursuc, N.~Komodakis, P.~P{\'e}rez, and M.~Cord, ``Boosting
  few-shot visual learning with self-supervision,'' in \emph{Proceedings of the
  IEEE/CVF international conference on computer vision}, 2019, pp. 8059--8068.

\bibitem{bib49}
B.~Liu, Y.~Cao, Y.~Lin, Q.~Li, Z.~Zhang, M.~Long, and H.~Hu, ``Negative margin
  matters: Understanding margin in few-shot classification,'' in \emph{Computer
  Vision--ECCV 2020: 16th European Conference, Glasgow, UK, August 23--28,
  2020, Proceedings, Part IV 16}.\hskip 1em plus 0.5em minus 0.4em\relax
  Springer, 2020, pp. 438--455.

\bibitem{bib50}
A.~Afrasiyabi, J.-F. Lalonde, and C.~Gagn{\'e}, ``Mixture-based feature space
  learning for few-shot image classification,'' in \emph{Proceedings of the
  IEEE/CVF international conference on computer vision}, 2021, pp. 9041--9051.

\bibitem{bib48}
H.-J. Ye, H.~Hu, D.-C. Zhan, and F.~Sha, ``Few-shot learning via embedding
  adaptation with set-to-set functions,'' in \emph{Proceedings of the IEEE/CVF
  conference on computer vision and pattern recognition}, 2020, pp. 8808--8817.

\bibitem{bib40}
B.~Liu, Y.~Cao, Y.~Lin, Q.~Li, Z.~Zhang, M.~Long, and H.~Hu, ``Negative margin
  matters: Understanding margin in few-shot classification,'' in \emph{Computer
  Vision--ECCV 2020: 16th European Conference, Glasgow, UK, August 23--28,
  2020, Proceedings, Part IV 16}.\hskip 1em plus 0.5em minus 0.4em\relax
  Springer, 2020, pp. 438--455.

\bibitem{bib45}
Y.~Tian, Y.~Wang, D.~Krishnan, J.~B. Tenenbaum, and P.~Isola, ``Rethinking
  few-shot image classification: a good embedding is all you need?'' in
  \emph{Computer Vision--ECCV 2020: 16th European Conference, Glasgow, UK,
  August 23--28, 2020, Proceedings, Part XIV 16}.\hskip 1em plus 0.5em minus
  0.4em\relax Springer, 2020, pp. 266--282.

\bibitem{bib43}
D.~Wertheimer, L.~Tang, and B.~Hariharan, ``Few-shot classification with
  feature map reconstruction networks,'' in \emph{Proceedings of the IEEE/CVF
  conference on computer vision and pattern recognition}, 2021, pp. 8012--8021.

\bibitem{bib41}
A.~Afrasiyabi, J.-F. Lalonde, and C.~Gagn{\'e}, ``Mixture-based feature space
  learning for few-shot image classification,'' in \emph{Proceedings of the
  IEEE/CVF international conference on computer vision}, 2021, pp. 9041--9051.

\bibitem{bib42}
Y.~Chen, Z.~Liu, H.~Xu, T.~Darrell, and X.~Wang, ``Meta-baseline: Exploring
  simple meta-learning for few-shot learning,'' in \emph{Proceedings of the
  IEEE/CVF international conference on computer vision}, 2021, pp. 9062--9071.

\bibitem{bib52}
J.~Atanbori and S.~Rose, ``Mergednet: A simple approach for one-shot learning
  in siamese networks based on similarity layers,'' \emph{Neurocomputing}, vol.
  509, pp. 1--10, 2022.

\bibitem{bib67}
G.~Wang, Y.~Wang, Z.~Pan, X.~Wang, J.~Zhang, and J.~Pan, ``Vitfsl-baseline: A
  simple baseline of vision transformer network for few-shot image
  classification,'' \emph{IEEE Access}, 2024.

\bibitem{bib44}
X.~Wang, X.~Wang, B.~Jiang, and B.~Luo, ``Few-shot learning meets transformer:
  Unified query-support transformers for few-shot classification,'' \emph{IEEE
  Transactions on Circuits and Systems for Video Technology}, vol.~33, no.~12,
  pp. 7789--7802, 2023.

\bibitem{bib51}
K.~Zheng, H.~Zhang, and W.~Huang, ``Diffkendall: a novel approach for few-shot
  learning with differentiable kendall's rank correlation,'' \emph{Advances in
  Neural Information Processing Systems}, vol.~36, pp. 49\,403--49\,415, 2023.

\bibitem{bib70}
Y.~Zhou, J.~Hao, S.~Huo, B.~Wang, L.~Ge, and S.-Y. Kung, ``Automatic metric
  search for few-shot learning,'' \emph{IEEE Transactions on Neural Networks
  and Learning Systems}, 2023.

\bibitem{bib54}
B.~Oreshkin, P.~Rodr{\'\i}guez~L{\'o}pez, and A.~Lacoste, ``Tadam: Task
  dependent adaptive metric for improved few-shot learning,'' \emph{Advances in
  neural information processing systems}, vol.~31, 2018.

\bibitem{bib53}
W.-Y. Chen, Y.-C. Liu, Z.~Kira, Y.-C.~F. Wang, and J.-B. Huang, ``A closer look
  at few-shot classification,'' \emph{arXiv preprint arXiv:1904.04232}, 2019.

\bibitem{bib55}
Y.~Wang, W.-L. Chao, K.~Q. Weinberger, and L.~Van Der~Maaten, ``Simpleshot:
  Revisiting nearest-neighbor classification for few-shot learning,''
  \emph{arXiv preprint arXiv:1911.04623}, 2019.

\bibitem{bib56}
K.~Lee, S.~Maji, A.~Ravichandran, and S.~Soatto, ``Meta-learning with
  differentiable convex optimization,'' in \emph{Proceedings of the IEEE/CVF
  conference on computer vision and pattern recognition}, 2019, pp.
  10\,657--10\,665.

\bibitem{bib57}
Y.~Lifchitz, Y.~Avrithis, S.~Picard, and A.~Bursuc, ``Dense classification and
  implanting for few-shot learning,'' in \emph{Proceedings of the IEEE/CVF
  conference on computer vision and pattern recognition}, 2019, pp. 9258--9267.

\bibitem{bib60}
Y.~Tian, Y.~Wang, D.~Krishnan, J.~B. Tenenbaum, and P.~Isola, ``Rethinking
  few-shot image classification: a good embedding is all you need?'' in
  \emph{Computer Vision--ECCV 2020: 16th European Conference, Glasgow, UK,
  August 23--28, 2020, Proceedings, Part XIV 16}.\hskip 1em plus 0.5em minus
  0.4em\relax Springer, 2020, pp. 266--282.

\bibitem{bib58}
F.~Gao, L.~Cai, Z.~Yang, S.~Song, and C.~Wu, ``Multi-distance metric network
  for few-shot learning,'' \emph{International Journal of Machine Learning and
  Cybernetics}, vol.~13, no.~9, pp. 2495--2506, 2022.

\bibitem{bib59}
H.~Chen, H.~Li, Y.~Li, and C.~Chen, ``Sparse spatial transformers for few-shot
  learning,'' \emph{Science China Information Sciences}, vol.~66, no.~11, p.
  210102, 2023.

\bibitem{bib65}
D.~Chakravarthi~Padmanabhan, S.~Gowda, E.~Arani, and B.~Zonooz, ``Lsfsl:
  Leveraging shape information in few-shot learning,'' \emph{arXiv e-prints},
  pp. arXiv--2304, 2023.

\bibitem{bib66}
Z.~Song, W.~Qiang, C.~Zheng, F.~Sun, and H.~Xiong, ``On the discriminability of
  self-supervised representation learning,'' \emph{arXiv preprint
  arXiv:2407.13541}, 2024.

\bibitem{bib72}
A.~Raghu, M.~Raghu, S.~Bengio, and O.~Vinyals, ``Rapid learning or feature
  reuse? towards understanding the effectiveness of maml,'' \emph{arXiv
  preprint arXiv:1909.09157}, 2019.

\bibitem{bib76}
J.~Oh, H.~Yoo, C.~Kim, and S.-Y. Yun, ``Boil: Towards representation change for
  few-shot learning,'' \emph{arXiv preprint arXiv:2008.08882}, 2020.

\bibitem{bib77}
J.~Von~Oswald, D.~Zhao, S.~Kobayashi, S.~Schug, M.~Caccia, N.~Zucchet, and
  J.~Sacramento, ``Learning where to learn: Gradient sparsity in meta and
  continual learning,'' \emph{Advances in Neural Information Processing
  Systems}, vol.~34, pp. 5250--5263, 2021.

\bibitem{bib78}
S.~Kang, D.~Hwang, M.~Eo, T.~Kim, and W.~Rhee, ``Meta-learning with a
  geometry-adaptive preconditioner,'' in \emph{Proceedings of the IEEE/CVF
  conference on computer vision and pattern recognition}, 2023, pp.
  16\,080--16\,090.

\bibitem{bib73}
S.~Bai, W.~Zhou, Z.~Luan, D.~Wang, and B.~Chen, ``Improving cross-domain
  few-shot classification with multilayer perceptron,'' in \emph{ICASSP
  2024-2024 IEEE International Conference on Acoustics, Speech and Signal
  Processing (ICASSP)}.\hskip 1em plus 0.5em minus 0.4em\relax IEEE, 2024, pp.
  5250--5254.

\bibitem{bib74}
W.~Li, L.~Wang, J.~Xu, J.~Huo, Y.~Gao, and J.~Luo, ``Revisiting local
  descriptor based image-to-class measure for few-shot learning,'' in
  \emph{Proceedings of the IEEE/CVF conference on computer vision and pattern
  recognition}, 2019, pp. 7260--7268.

\bibitem{bib75}
R.~Hou, H.~Chang, B.~Ma, S.~Shan, and X.~Chen, ``Cross attention network for
  few-shot classification,'' \emph{Advances in neural information processing
  systems}, vol.~32, 2019.

\bibitem{bolhasani2020histopathological}
H.~Bolhasani, E.~Amjadi, M.~Tabatabaeian, and S.~J. Jassbi, ``A
  histopathological image dataset for grading breast invasive ductal
  carcinomas,'' \emph{Informatics in Medicine Unlocked}, vol.~19, p. 100341,
  2020.

\end{thebibliography}

\end{document}